%% file: main.tex
\documentclass[10pt,twocolumn,letterpaper]{article}

\usepackage{iccv}
\usepackage{times}
\usepackage{epsfig}
\usepackage{graphicx}
\usepackage{amsmath}
\usepackage{amssymb}
\usepackage{tabularx}
\usepackage{array}
\usepackage{booktabs}
\usepackage{algorithm}
\usepackage{algpseudocode}
\usepackage{multirow}
\usepackage{xcolor}
\usepackage{cuted}

\newcommand{\pointcloud}{\mathcal X}
\newcommand{\pointCount}{N}
\newcommand{\pointCountClean}{M}
\newcommand{\point}{\mathbf x}
\newcommand{\noisyPointcloud}{\mathcal Y}
\newcommand{\noisyPoint}{\mathbf y}
\newcommand{\position}{\mathbf z}
\newcommand{\noiseModel}{p}
\newcommand{\surfaceSample}{\mathbf s}
\newcommand{\surface}{\mathcal S}
\newcommand{\mapping}{f}
\newcommand{\tunableParameters}{\Theta}
\newcommand{\loss}{l}
\newcommand{\prior}{q}
\newcommand{\priorSample}{\mathbf q}
\newcommand{\priorSamples}{\mathcal Q}
\newcommand{\randomSeed}{\xi}
\newcommand{\randomSeeds}{\Xi}

\newcommand{\proximityMatrix}{\mathsf W}
\newcommand{\proximityMatrixElement}{w}
\newcommand{\pathcRadius}{r}
\newcommand{\pathcRadiusScale}{\alpha}
\newcommand{\matElementAppear}{\beta}
\newcommand{\kernel}{k}
\newcommand{\offset}{\mathbf d}

\usepackage{pifont}

\newcolumntype{C}[1]{>{\centering\arraybackslash}m{#1}}
\newcolumntype{R}[1]{>{\raggedleft\arraybackslash}m{#1}}

\usepackage[pagebackref=true,breaklinks=true,letterpaper=true,colorlinks,bookmarks=false]{hyperref}

\definecolor{MeanColor}{rgb}{0.556,0.42,0.9}
\definecolor{BilateralColor}{rgb}{0.12,0.45,0.76}
\definecolor{NoPriorColor}{rgb}{0.274,0.74,0.77}
\definecolor{NoColColor}{rgb}{0.26,0.78,0.19}
\definecolor{FullColor}{rgb}{1,0.6,0}
\definecolor{SupervisedColor}{rgb}{1,0,0}

\input{tr-commands.tex}

\iccvfinalcopy

\def\iccvPaperID{5698}

\ificcvfinal\fi
\begin{document}

\title{Total Denoising: Unsupervised Learning of 3D Point Cloud Cleaning}

\author{Pedro Hermosilla\\
Ulm University
\and
Tobias Ritschel\\
University College London
\and
Timo Ropinski\\
Ulm University
}

\maketitle

\begin{abstract}
We show that denoising of 3D point clouds can be learned unsupervised, directly from noisy 3D point cloud data only. This is achieved by extending recent ideas from learning of unsupervised image denoisers to unstructured 3D point clouds. Unsupervised image denoisers operate under the assumption that a noisy pixel observation is a random realization of a distribution around a clean pixel value, which allows appropriate learning on this distribution to eventually converge to the correct value. Regrettably, this assumption is not valid for unstructured points: 3D point clouds are subject to \emph{total noise}, \ie deviations in all coordinates, with no reliable pixel grid. Thus, an observation can be the realization of an entire manifold of clean 3D points, which makes a na\"ive extension of unsupervised image denoisers to 3D point clouds impractical. Overcoming this, we introduce a spatial prior term, that steers converges to the unique closest out of the many possible modes on a manifold. Our results demonstrate unsupervised denoising performance similar to that of supervised learning with clean data when given enough training examples - whereby we do not need any pairs of noisy and clean training data.
\end{abstract}

\mysection{Introduction}{Introduction}
While the amount of clean 3D geometry is limited by the manual effort of human 3D CAD modelling, the amount of 3D point clouds is growing rapidly everyday: our city's streets, the interior of everyday buildings, and even the goods we consume are routinely 3D-scanned. Regrettably, these data are corrupted by scanner noise and as such not accessible to supervised learning that requires pairs of noisy and clean data. Consequently, it is desirable to be able to denoise the acquired noisy 3D point clouds by solely using the noisy data itself.

Two necessary recent developments indicate that this might be possible: deep learning on 3D point clouds~\cite{qi2016pointnet} and unsupervised denoising of images~\cite{lehtinen2018noise2noise,krull2018noise2void,batson2019noise2self}.

\myfigure{Teaser}{We learn 3D point cloud cleaning \emph{(right)}, unsupervised, from noisy examples alone \emph{(left)}.}{t}

Unfortunately, these two methods cannot be combined na\"ively. To learn our unsupervised 3D point cloud denoisers (\refFig{Teaser}), we need to overcome two main limitations: the practical obstacle to have a \emph{pair} of two noisy scans of the same object and the theoretical difficulty that noise in 3D point clouds is \emph{total}.

We refer to noise as `total' (\refFig{TotalNoise}) when distortions are not confined to the range (pixel values) while the domain is clean (as pixel positions are), but to the more challenging setting where both domain and range are affected by noise. The name is chosen in analogy to total least squares~\cite{golub1980analysis}, which dealt with simultaneous noise in domain and range, but for a linear, non-deep setting.

\myfigure{TotalNoise}{
To learn denoising of 3D point clouds, we need to extend from common noise that is clean in one part of the signal, to a total setting, where all parts of the signal are noisy. This example shows three realizations of common noise \emph{(left)} and total noise \emph{(right)} for three samples \emph{(colors)}. Please note, how total noise is ``more noisy'' as both axis are corrupted.}{b}

This paper's evaluation shows for simulated noise of different kind, as well as for real point clouds, how our unsupervised approach nonetheless outperforms a supervised approach given enough, and in some cases even when given the same magnitude of, training data, while it runs efficiently in a single pass on large point clouds.

\mysection{Related Work}{RelatedWork}

\paragraph{Image denoising.}
Denoising images is one of the most basic image manipulation operations.
The most primitive variants are based on linear filters such as the Gaussian filter, eventually with additional sharpening~\cite{jain1989fundamentals}. While non-linear filters, such as median, bilateral \cite{tomasi1998bilateral} or non-local means~\cite{buades2005non} are frequently used in practice, state-of-the-art results are achieved by optimizing for sparsity~\cite{elad2006image}. Recently, it has become popular to learn denoising, when pairs of clean and noisy images are available~\cite{burger2012image}.

Lehtinen~\etal\cite{lehtinen2018noise2noise} proposed a method to learn denoising with access to only two noisy images, instead of a clean-noisy pair. Taking it a step further, Noise2Void~\cite{krull2018noise2void} and Noise2Self~\cite{batson2019noise2self} are two extensions that remove the requirement to have two copies of one image corrupted with noise and instead work on a single image. In both cases, this is achieved by regressing the image from itself. This is done by creating a receptive field with a ``blind spot'', and a network regresses the blind spot from its context. We will detail the theory behind those papers~\cite{lehtinen2018noise2noise,krull2018noise2void,batson2019noise2self} in \refSec{Background}.

\mycfigure{StructuredVsUnstructured}{Substantial differences exist when denoising structured and unstructured data. \emph{(a)} For structured data, each pixel value follows a sampling distribution $\noiseModel(\position|\point_i)$ (\emph{yellow curve}) around the true value (pink circle). \emph{(b)} For unstructured data, the distribution $p(\position|\surface)$ has a manifold of modes (\emph{pink line}). \emph{(c)} By using the proposed proximity-appearance prior, a unique mode closest to the surface is determined.}

\paragraph{3D point cloud denoising.}
3D point clouds capture fine spatial details but remain substantially more difficult to handle than images, due to their irregular structure~\cite{levoy1985use}.

As for images, linear filters can be applied to remove noise~\cite{lee2000curve}, but at the expense of details. As a remedy, image operators such as bilateral~\cite{fleishman2003bilateral, Digne2017Bilateral}, non-local means~\cite{rosman2013patch} or sparse coding~\cite{avron2010l1} have been transferred to point clouds.

With the advent of PointNet~\cite{qi2016pointnet}, deep learning-based processing of point clouds has become tractable.
Four notable deep methods to denoise 3D point clouds were suggested.
The first is PointProNet, that denoises patches of points by projecting them to a learned local frame and using Convolutional Neural Networks (CNN) in a supervised setup to move the points back to the surface~\cite{roveriO2018pointpronet}.
However, the accuracy of the method is determined by the accuracy of the local frame estimation, which results in artifacts in extreme sharp edges.
The second approach by Rakotosaona \etal~\cite{rakotosaona2019pointclean} uses PCPNet~\cite{Guerrero2017PCPNet} (a variant of PointNet~\cite{qi2016pointnet}) to map noisy point clouds to clean ones.
Third, Yu~\etal\cite{yu2018ec} learns to preserve edges, that dominate men-made objects.
Finally, Yifan~\etal\cite{yifan2019patch} define a clean surface from noisy points by upsampling.

All these deep denoising approaches are supervised, as they require pairs of clean and noisy point clouds, which in practice are produced by adding noise to synthetic point clouds.
Our approach does not require such pairs.

\paragraph{Noise and learning.}
Noise is an augmentation strategy used in denoising auto-encoders~\cite{vincent2010stacked}. These are however not aiming to denoise, but add noise to improve robustness. Also is their target not noisy, but noise is in the input or added to internal states.

\mysection{Denoising Theory}{Background}
Based on denoising in the regular domain, \ie images, with or without supervision, we will establish a formalism that can later also be applied to derive our unstructured 3D case.

\mysubsection{Regular Domains}{RegularNoise}

\paragraph{Pixel noise.}
An observation $\noisyPoint_i$ at pixel $i$ in a noise corrupted image is a sample of a noise distribution $\noisyPoint_i\sim\noiseModel(\position|\point_i)$ around the true value $\point_i$. This is shown in \refFig{StructuredVsUnstructured}, a). The black curve is the true signal and pixels (dotted vertical lines) sample it at fixed positions $i$ (black circles) according to a sampling distribution $\noiseModel(\position|\point_i)$ (yellow curve) around the true value (pink circle).

\paragraph{Supervised.}
In classic supervised denoising, we know both a clean $\point_i$ and a noisy value $\noisyPoint\sim\noiseModel(\position|\point_i)$ for pixel $i$ and minimize
\[
\argmin\tunableParameters
\
\mathrm E_{
 \noisyPoint
 \sim
 \noiseModel(\position|\point_i)
}
\loss(
 \mapping_\tunableParameters(\noisyPoint),
 \point_i),
\] where $f$ is a tunable function with parameters $\tunableParameters$, and $\loss$ is a loss such as $L_2$. Here and in the following, we omit the fact that the input to $\mapping$ comprises of many $\noisyPoint$ that form an entire image, or at least a patch. We also do not show an outer summation over all images (and later, point cloud) exemplars.

\paragraph{Unsupervised, paired.}
Learning a mapping from one noisy realization of an image to another noisy realization of the same image is achieved by Noise2Noise~\cite{lehtinen2018noise2noise}.
It has been shown, that learning \[
\argmin\tunableParameters
\
\mathrm E_{
 \noisyPoint_1
 \sim
 \noiseModel(\position|\point_i)}
\mathrm E_{
 \noisyPoint_2
 \sim
 \noiseModel(\position|\point_i)
}
\loss(
 \mapping_\tunableParameters(\noisyPoint_1),
 \noisyPoint_2),
\]
converges to the same value as if it had been learned using the mean / median / mode of the distribution $p(\position|\point)$ when $\loss$ is $L_2$ /  $L_1$ / $L_0$. In most cases, \ie for mean-free noise, the mean / median / mode is also the clean value. We refer to this method as `paired', as it needs two realizations of the signal, \ie one image with two realizations of noise.

\paragraph{Unsupervised, unpaired.}
Learning a mapping from all noisy observations in one image, except one pixel, to this held-out pixel is achieved by Noise2Void~\cite{krull2018noise2void} and Noise2Self~\cite{batson2019noise2self}: 
\[
\argmin\tunableParameters
\
\mathrm E_{
 \noisyPoint
 \sim
 \noiseModel(\position|\point_i)
}
\loss(
 \mapping_\tunableParameters(\noisyPoint),
 \noisyPoint),
\]
Here, $\mapping$ is a special form of $\mathcal J$-incomplete~\cite{batson2019noise2self} maps that have no access to pixel $i$ when regressing it, \ie a `blind spot'. The same relation between mean / median / mode and loss as in Noise2Noise applies. Note that this formulation does not require two images, and we, therefore, refer to it as `unpaired'.

\paragraph{Discussion.}
All three methods described above, work under the assumption that, in a structured pixel grid, the range (vertical axis in \refFig{TotalNoise}, left and \refFig{StructuredVsUnstructured}, a) axis $i$ and the domain $\position$ (horizontal axis) have different semantics. The noise is only in the range: it is not uncertain \emph{where} a pixel is, only what its true \emph{value} would be.

\mysubsection{Unstructured Domains}{UnstructuredNoise}

\paragraph{Point noise.}
As for pixels, we will denote clean points as $\point$, noisy points as $\noisyPoint$ and the noise model as $\noiseModel$. All points in our derivation can be either positional with XYZ coordinates, or positional with appearance, represented as XYZRGB points.

To our knowledge, deep denoising of colored point clouds has not been proposed. We will not only show how our technique can also be applied to such data but moreover, how color can help substantially to overcome challenges when training unsupervised learning of a point cloud denoiser. Surprisingly, this benefit can be exploited during training, even when no color is present at test time. If available, it will help, and we can also denoise position and appearance jointly.

\paragraph{Supervised.}
Denoising a point cloud means to learn
\[
\argmin\tunableParameters
\
\mathrm E_{
 \noisyPoint
 \sim
 \noiseModel(\position|\surface)
}
\loss(
 \mapping_\tunableParameters(\noisyPoint),
 \surface),
\]
the sum of the losses $\loss$ (\eg Chamfer) between $\mapping_\tunableParameters(\noisyPoint)$ and the surface $\surface$ of the 3D object. Such supervised methods have been proposed, but they remain limited by the amount of training data available~\cite{roveriO2018pointpronet,rakotosaona2019pointclean}, as they require access to a clean point cloud.

\mysection{Unsupervised 3D Point Cloud Denoising}{OurApproach}
We will first describe why a paired approach is not feasible for unstructured data before we introduce our unpaired, unsupervised approach.

\mysubsection{Inapplicability of `Paired' Approaches}{Paired}
Learning a mapping $\mapping_\tunableParameters(\noisyPointcloud_1)=\noisyPointcloud_2$ from one noisy point cloud realization $\noisyPointcloud_1$ to another noisy point cloud realization $\noisyPointcloud_2$ that both have the same clean point cloud $\pointcloud$ and where the $i$-th point in both point clouds is a realization of the $i$-th ground truth value, would be a denoiser in the sense of Noise2Noise~\cite{lehtinen2018noise2noise}. Regrettably, Noise2Noise cannot be applied to unsupervised learning from unstructured point clouds for two reasons.

First, this paired design, same as for images, would require supervision in the form of two realizations of the same point cloud corrupted by different noise realizations. While this is already difficult to achieve for 2D image sensors, it is not feasible for 3D scanners.

Second, it would require a network architecture to know which point is which, similar as it is given by the regular structure of an image that explicitly encodes each pixel's identity $i$. This is never the case for total noise in points. Opposed to this, modern convolutional deep point processing~\cite{qi2016pointnet,hermosilla2018mccnn} is exactly about becoming invariant under reordering of points.

In order to overcome this problem in a supervised setting, Rakotosaona \etal~\cite{rakotosaona2019pointclean} simulated such pairing by selecting, for each noisy observation, the closest point in the clean point cloud as the target for the loss. However, this is just an approximation of the real surface whose accuracy depends on the quality of the sampling of the clean data. Fortunately, we can show that a pairing assumption is not required, such that our approach operates not only unsupervised but also unpaired, as we will detail next.

\mysubsection{Unpaired}{Unpaired}
Learning a mapping from a noisy realization to itself $\mapping_\tunableParameters(\noisyPointcloud)=\noisyPointcloud$ is an unsupervised and unpaired denoiser in the sense of Noise2Void~\cite{krull2018noise2void} or Noise2Self~\cite{batson2019noise2self}. Defining $\mathcal J$ incompleteness in a point cloud is no difficulty: just prevent access of $f$ to point $\noisyPoint$ itself when learning point $\noisyPoint$ from the neighbors of $\noisyPoint$. Thus, essentially, we train a network to map each point to itself without information about itself. Unfortunately, there is the following catch with total noise.

\paragraph{Problem statement.}
Different from observing pixels at index $i$  in an image (dotted line \refFig{StructuredVsUnstructured}, a), which tell us that $\noisyPoint$ is a realization of a hidden value $\point_i$ to infer, it is unknown which hidden surface point is realized when observing a point in an unpaired setting. A noisy point observation $\noisyPoint$, can be a realization of $\noiseModel(\position|\point_1)$ in the same way as it could be a realization of $\noiseModel(\position|\point_2)$. Consequently, the distribution $p(\position|\surface)$ has a manifold of modes (pink line in \refFig{StructuredVsUnstructured}, b). Learning a mapping from a noisy realization to itself will try to converge to this multimodal distribution, since, for the same neighborhood, the network will try to regress different points from this distribution at the same time.

We, therefore, have to look into two questions. First, what can be said about the similarity of this manifold of modes and the clean surface? And second, how can we decide which of the many possible modes is the right one? Answering the second, and deriving bounds for the first question are the key contributions of this paper, enabling unsupervised 3D point cloud denoising.

\mysubsection{Manifold of Modes vs. Clean Surface}{ManifoldVsClean}
Concerning the first question, the manifold of modes is close to the surface, but not identical. \refFig{ManifoldVsSurface}, a), shows a clean surface as a black line, with a small amount of noise, where most samples are close to the clean surface. In this condition, the learning converges to a solution identical to a solution it would have converged to, as when trained on the pink line, which is very similar to the clean surface. With more noise, however, it becomes visible in \refFig{ManifoldVsSurface}, b) that this manifold is not identical to the surface.

\myfigure{ManifoldVsSurface}{Comparing small (\emph{left}) and large noise (\emph{right}) we see the modes (\emph{pink}) deviate from the GT surface (\emph{black}).}{t}

We note, that the mode surface is the convolution of the true surface and the noise model $\noiseModel$. We cannot recover details removed by this convolution. This is different from supervised NN-based deconvolution, which has access to pairs of convolved and clean data. In our case, the convolution is on the limit case of the learning data and we never observe non-convolved, clean data.

It is further worth noting, that not all noise distributions lead to a manifold that is a surface in 3D or would be a connected path in our 2D illustrations. Only uni-modal noise distributions, such a scanner noise, have no branching  or disconnected components. 
Our solution will not depend on the topology of this mode structure.

\mysubsection{Unique Modes}{UniqueModes}
As explained above, the na\"ive implementation of unsupervised unpaired denoising will not have a unique mode to converge to. Therefore, we regularize the problem by imposing the prior $\prior(\position|\noisyPoint)$ that captures the probability that a given observation $\noisyPoint$ is a realization of the clean point $\position$. 

We suggest using a combination of spatial and appearance proximity \begin{align}
\prior(\position|\noisyPoint)
&=
\noiseModel(\position|\surface)*\kernel(\position-\noisyPoint)
\\
\kernel(\offset)&= \frac{1}{\sigma\sqrt{2\pi}}\exp\left(-\frac{||\proximityMatrix\offset||_2^2}{2\sigma^2}\right),
\end{align}
where $\sigma$ is the bandwidth of $\kernel$ and $\proximityMatrix=\operatorname{diag}(\proximityMatrixElement)$ is a diagonal weight matrix trading spatial and appearance locality. We use a value $\proximityMatrixElement=1/\pathcRadiusScale\pathcRadius$, $\pathcRadius$ being $5\%$ of the diameter of the model and $\pathcRadiusScale$ a scaling factor. In the case of point clouds with appearance, we use $\proximityMatrixElement=\matElementAppear$ in the appearance rows/columns, otherwise, we only consider proximity. For more details about the values for such parameters please refer to the supplementary material.

This results in convergence to the nearest (in space and appearance) mode when optimizing \begin{align}
\label{eq:Our}
\argmin\tunableParameters
\
\mathrm E_{
 \noisyPoint
 \sim
 \noiseModel(\position|\surface)
}
\mathrm E_{
 \priorSample
 \sim
 \prior(\position|\noisyPoint)
}
\loss(
 \mapping_\tunableParameters(\noisyPoint),
 \priorSample),
\end{align}

The effect of this prior is seen in \refFig{StructuredVsUnstructured}, c). Out of many modes, the unique closest one remains.

Note, that our choice of a Gaussian prior $\prior$ is not related to a Gaussianity of the noise model $\noiseModel$, which we do not assume. The only assumption made here is that out of many explanations, the closest one is correct. We experimented with other kernels such as Wendland~\cite{wendland1995piecewise} and inverse multi-quadratic but did not observe an improvement.


\paragraph{Appearance to the rescue}
\myfigure{Bilateral}{Bilaterality:
The manifold of modes of the distribution of a 2D point cloud without color can be curved for strong noise (\emph{a}). Different appearancea, denoted as red and blue points in \emph{b}, can be used to establish bilateral distances, lifting points to 3D. The resulting manifold of modes (\emph{d}) now preserves sharp appearance edges.
}{t}

As mentioned above, 3D point clouds that come with RGB color annotation are a surprising opportunity to further overcome the limitations of unsupervised training. Otherwise, in some cases, the spatial prior cannot resolve round edges. This is not because the network $\mapping$ is unable to resolve them, but because unsupervised training does not `see' the sharp details. \refFig{Bilateral} details how colors resolve this: without RGB, the corners are rounded in \refFig{Bilateral}, a). When adding color, here red and blue (\refFig{Bilateral}. b), the points become separated (\refFig{Bilateral}, c). The sampling of the prior $\prior(\position|\noisyPoint)$ on a red point, will never pick a blue one and vice-versa. Consequently, the learning behaves as if it had seen the sharp detail.

Thus, using color in the prior reinforces some of the structure, which was lost when not relying on a regular pixel grid. We do not know which noisy point belongs to which measurement, but we have a strong indication, that something of different color, is not a noisy observation of the same point. Of course, it is possible, that two observations $\noisyPoint_1$ and $\noisyPoint_2$ appear to be from a different point, but happen to be measurements of the same clean surface point $\point$, whereby range noise affects the color. Fortunately, such spurious false negatives are less problematic (they create variance) than the permanent false positives, that lead to rounding (bias). Symmetrically, and maybe more severe, a difference in color is not always a reliable indicator of a geometric discontinuity either. It is, if color is dominated by shading, but texture and shadow edges may lead to a high false negative distance. Note, that the color is only required for training and never used as input to the network.

\mysubsection{Converging to the mode}{ConvergingMode}
We train the network to converge to the mode of the prior distribution $\prior(\position|\noisyPoint)$ by using the approximation of the $L_0$ loss function proposed by Lehtinen et al.~\cite{lehtinen2018noise2noise}, $( | \mapping_\tunableParameters(\noisyPoint) - \priorSample | + \epsilon)^\gamma$, 
where their $\epsilon=10^{-8}$, and their $\gamma$ is annealed from $2$ to $0$ over the training.

Thus, our unsupervised training converges to the same as training supervised to find the closest -- in XYZ space and RGB appearance, when available --  mode on the distribution  $\surface*\noiseModel$ resulting from convolving the clean surface $\surface$ and the noise model $\noiseModel$.

\mysubsection{Implementation}{Implementation}

\paragraph{Prior.}
To minimize \refEq{Our} we need to draw samples according to the prior $\prior$ which is implemented using rejection sampling: we pick a random point $\hat\priorSample$ from $\noisyPointcloud$ within $\pathcRadius$ from $\noisyPoint$, and train on it only if $\kernel(\hat\priorSample - \noisyPoint)>\randomSeed$ for a uniform random $\randomSeed\in(0,1)$. In practice, a single sample is used to estimate this inner expected value over $\prior(\position| \noisyPoint)$.

\paragraph{Architecture.}

\myfigure{Architecture}{Architecture overview:
We start from a noisy point cloud in the top and perform two levels of unstructured encoding, that reduce the receptive field, followed by two levels of decoding using transposed unstructured convolutions.}{t}

We implement $\mapping$ using an unstructured encoder-decoder based on Monte Carlo convolution~\cite{hermosilla2018mccnn} (\refFig{Architecture}). Such an architecture consumes the point cloud, transforms spatial neighborhoods into latent codes defined on a coarser point set (encoder), and up-sample these to the original point resolution (decoder). The effective receptive field, so the neighborhood from which the NN regressed points are considered, is $30\,\%$ of the diameter of the model. In particular, we perform two levels of encoding, the first with a receptive field of $5\,\%$, the second at $10\,\%$. The Poisson disk radii for pooling in Level 1 and Level 2 are half the size of the receptive fields. 

This architecture is fast to execute, allowing to denoise in parallel $800\,$K points in $13$ seconds on a single machine with a GeForce RTX 2080. Moreover, this architecture is composed of only $25\,K$ trainable parameters, orders of magnitude smaller than other networks ($0.8\,$million for PointNet or $1.4\,$million for PointNet++).

\paragraph{Training.}
Besides these benefits, our method is also easy to implement as seen in~\refAlg{Ours}. Here, $\priorSamples$ denotes a set of prior samples $\priorSample$ for all points in a point cloud. All operations are defined on batches that have the size of the point cloud. We use an ADAM optimizer~\cite{KingmaB14Adam} with an initial learning rate of $.005$, which is decreased during training.

\begin{algorithm}[htb]
    \caption{Unsupervised point cloud denoiser training}
    \label{alg:Ours}
    \begin{algorithmic}[1]
        \For{all noisy point clouds $\noisyPointcloud$}
        
        \State $\randomSeeds \gets \Call{RandomUniformBatch}{0,1}$
        
        \State $\priorSamples 
        \gets 
        \Call{samplePriorBatch}{\noisyPointcloud, \randomSeeds}$
        
        \State $\tunableParameters
        \gets
        \Call{minimizeBatch}{||                   
                \mapping_\tunableParameters(\noisyPointcloud)- \priorSamples||_0}$
        \EndFor
    \end{algorithmic}
\end{algorithm}

\paragraph{Iteration.}
Similar as in previous work~\cite{rakotosaona2019pointclean}, our results improved if the output of the network is fed as input again. However, this introduces two problems: clustering of points, and shrinking of the point cloud after several iterations. We address these problems in a similar way as Rakotosaona \etal~\cite{rakotosaona2019pointclean}. In order to prevent clustering we introduce the following regularization term that enforces a point cloud with equidistant samples:
\[
L_{r}=\argmin\tunableParameters
\
\mathrm E_{
 \noisyPoint
 \sim
 \noiseModel(\position|\surface)
}
\max_{
 \noisyPoint^{\prime}
 \in
 n(\noisyPointcloud, \noisyPoint)
}
\|\mapping_\tunableParameters(\noisyPoint), \mapping_\tunableParameters(\noisyPoint^{\prime})\|_2
\]
\noindent where $n(\noisyPointcloud, \noisyPoint)$ is the set of points from the noisy point cloud within a patch centered at $\noisyPoint$.
%
To prevent shrinking we remove low-frequency displacements before translating the noisy points. The supplemental materials show the effect of these iterations.

\mysection{Evaluation}{Evaluation}
Our experiments explore the application, both to synthetic (\refSec{Quantitative}) and to real data (\refSec{Qualitative}). For synthetic data, we know the answer and can apply different metrics to quantify the performance of our approach, while we do not know the ground truth for real data and results are limited to a qualitative study.

\mysubsection{Setup}{Setup}

\paragraph{Data set.}
We have collected $15$ different classes with $7$ different polygonal models each ($5$ for training and $2$ for testing) from ModelNet-40 \cite{Wu2015ModelNet40} and sampled the surface with points as explained next. As we optionally use RGB appearance, it is computed using Lambertian shading from 3 random directional lights.

\paragraph{Sampling.}
We simulate different forms of noise to corrupt the clean data of the synthetic data set.

In the \textsc{Simple} noise model, we sample each mesh using Poisson Disk sampling~\cite{Wei2008Poisson} to obtain clean point clouds within the range of $13\,$K and $190\,$K points each, resulting in $22\,$million points for training and $10\,$million points for testing. Then, we add Gaussian noise with a standard deviation of $.5\%$, $1\%$, and $1.5\%$ of the bounding box diagonal.

The \textsc{Advanced} sampling emulates true sensor noise making use of Blendsor~\cite{gschwandtner2011blensor}, a library to simulate sensor noise. 
In particular, we choose to emulate a Velodyne HDL-64E 3D scan.
These devices introduce two types of noise in the measurements, a distance bias for each laser unit and a per-ray Gaussian noise. 
In our data, we use a standard deviation of $.5\%$ of the diagonal of the bounding box for the distance bias and three different levels of per-ray noise, $.5\%$, $1\%$, and $1.5\%$.
This generates point clouds within the range of $3\,$K and $120\,$K points each, resulting in $12\,$million points for training and $5\,$million points for testing.

We investigate levels of distortion where a surface is still conceivable.
More severe corruptions with uneven sampling or outliers are to be explored in future work.

\paragraph{Metric.}
We use the Chamfer distance from Fan et al.~\cite{Fan2016GAN},

\[
d(\noisyPointcloud, \surface, \pointcloud) =
\frac{1}{\pointCount}
\sum_{\noisyPoint\in\noisyPointcloud}
\min_{\surfaceSample\in\surface}
||\noisyPoint-\surfaceSample||_2 
+
\frac{1}{\pointCountClean}
\sum_{\point\in\pointcloud}
\min_{\noisyPoint\in\noisyPointcloud}
||\noisyPoint-\point||_2
\]
\noindent where less is better. The first term measures the average distance between the predicted points to their closest point in a polygonal surface $\surface$. The second term measures how the points are distributed in the ground truth surface.

Since the clean point clouds $\pointcloud$ follow a Poisson Disk distribution, by measuring their distance to the closest predicted point we are able to determine if the surface is equally covered by our predictions.

The metric is applied in the test data with three different realizations of noise and averaged over two complete trainings to reduce variance in the estimate of the metric.

\paragraph{Methods.}
We compare our unsupervised approach with classical methods as well as supervised machine learning approaches. To obtain insights regarding the effectivity of the individual subparts, we investigate ablations with and without the spatial and/or the appearance prior.

The classic baselines are \textsc{Mean} and \textsc{Bilateral}~\cite{Digne2017Bilateral}, which are also unsupervised. Their parameters are chosen to be optimal on the training set.

As supervised learning-based denoisers, we use the same architecture, as we have employed for the unsupervised setting, whereby we use the training algorithm proposed in PointCleanNet~\cite{rakotosaona2019pointclean}. While this means, that we do not use the original PointCleanNet network architecture, which is based on PointNet, we believe that our evaluation is more insightful with a unified architecture -- especially since the architecture employed by us has outperformed PointNet on a range of other tasks~\cite{hermosilla2018mccnn}.

Finally, we study three variants of our approach. The first one is with \textsc{No Prior}. The second we denote as \textsc{No Color}, which is our prior but only based on proximity. The last one is \textsc{Full} which includes all our contributions. More precisely, we use XYZ point clouds for \textsc{No Prior} and \textsc{No Color} and XYZRGB point clouds for \textsc{Full}. Again, color is only used to sample the prior, not as an input to the network during training or testing.

\mysubsection{Quantitative Results}{Quantitative}


\paragraph{Denoising performance.}
 We start with \textsc{Simple} noise and look into \textsc{Advanced} later. A comparison of the average error across the test set for different methods is shown in \refTbl{Performance}, whereby each column represents one method. All methods are trained with the same amount of training exemplars, that is, $22\,$million points. 

As can be seen, our full method (orange) performs best. We even outperform the supervised competitor (red), likely because the network has to find more valid generalizations and is less prone to over-fitting. As can also be seen, the other non-learned methods like mean (violet) and bilateral (blue) are not competitive, even when tuned to be optimal on training data.
We further see a clear distinction between ablations of our method and the full approach. When training without a spatial prior (cyan), the method is much worse than supervised and only slightly better than mean. A method not using color for training (green) -- but including the spatial prior -- can achieve almost full performance, but only adding color will outperform supervised.

\begin{table}[t]
\caption{Error (less is better) per method on \textsc{Simple} noise.}
\label{tbl:Performance}
\begin{minipage}[b]{0.72\linewidth}
\setlength{\tabcolsep}{3.0pt}
\begin{tabular}{rrrrrr}
 & & \multicolumn{3}{c}{Ours} & \\
\cmidrule(lr{.75em}){3-5}
Mean&
Bilat.&
No p.&
No c.&
Full&
Sup.
\\
\multicolumn{1}{c}{\textcolor{MeanColor}{\textbullet}}&
\multicolumn{1}{c}{\textcolor{BilateralColor}{\textbullet}}&
\multicolumn{1}{c}{\textcolor{NoPriorColor}{\textbullet}}&
\multicolumn{1}{c}{\textcolor{NoColColor}{\textbullet}}&
\multicolumn{1}{c}{\textcolor{FullColor}{\textbullet}}&
\multicolumn{1}{c}{\textcolor{SupervisedColor}{\textbullet}}
\\
\toprule
.598 & .592 & .582 & .547 & .542 & .545\\
\bottomrule
\end{tabular}%
\end{minipage}%
\begin{minipage}[b]{0.28\linewidth}
\includegraphics[width=2cm, height=2cm]{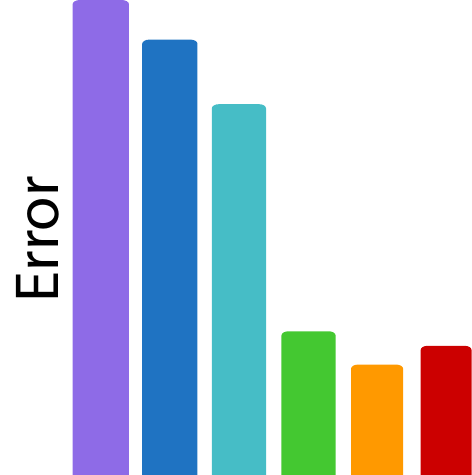}%
\vspace{-1.0cm}
\end{minipage}
\end{table}

\begin{table}[b]
\caption{Error (less is better) for different amount of supervision.}
\label{tbl:Scalability}
\begin{minipage}[b]{0.6\linewidth}
\setlength{\tabcolsep}{2.5pt}
\begin{tabular}{rrrrr}
& \multicolumn{3}{c}{Ours}\\
\cmidrule(lr{.75em}){2-4}
\multirow{2}{1 cm}{Train data}&
No p.&
No c.&
Full&
Sup.
\\
&
\multicolumn{1}{c}{\textcolor{NoPriorColor}{\textbullet}}&
\multicolumn{1}{c}{\textcolor{NoColColor}{\textbullet}}&
\multicolumn{1}{c}{\textcolor{FullColor}{\textbullet}}&
\multicolumn{1}{c}{\textcolor{SupervisedColor}{\textbullet}}
\\
\toprule
.5\,M & .587 & .557 &.558 &.574\\
1\,M & .584	& .550 &.557 &.563\\
4\,M & .584	& .553 &.543 &.546\\
22\,M & .582 & .547 &.542 &.545\\
\bottomrule
\end{tabular}%
\end{minipage}%
\begin{minipage}[b]{0.3\linewidth}
\includegraphics[width=3.25cm]{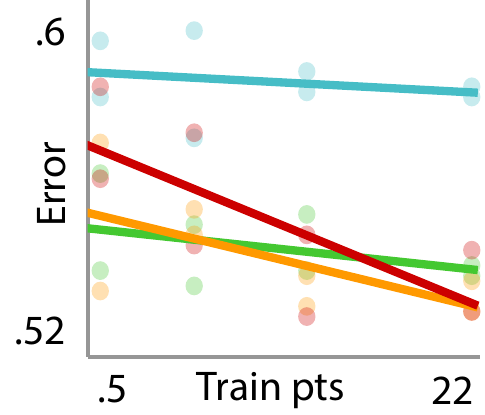}%
\vspace{-1.75cm}
\end{minipage}
\end{table}

This comparison is on the same amount of data. However, in most real-world scenarios, the number of noisy point clouds can be assumed to be much higher than the amount of clean ones. We will study this relation next.

\paragraph{Supervision scalability.}
We will now study, how supervised method scale with the number of clean point clouds and our method with the amount of noisy point clouds.

The outcome is seen in \refTbl{Scalability} where different methods are columns and different amounts of training data are rows. The plot to the right shows the relation as a graph. We show the logarithmic fit to the individual measurements shown as points
The color encoding is identical for all plots. The difference between the methods is measured wrt.\ the number of total training points ranging from $.5$ to $22\,$millions.

Not unexpected, we see all methods benefit from more training data. We see that our method performs better than supervised across a range of training data magnitudes. At around $22\,$million points, the highest we could measure, the red and orange lines of our full model and supervised cross. That only means, that after this point, our unsupervised method needs more training data to achieve the same performance as a supervised method.

We further see that the ablations of our method without a prior and without color do not only perform worse, but also scale less favorable, while ours (orange) is similar to supervised (red). Admittedly, supervised scales best.

\mycfigure{RealData}{Multiple real world pairs of noisy input scans \emph{(left)} and the result of our denoiser \emph{(right)}, accompanied by zoomed insets.}


\paragraph{Amount of noise.}

\begin{table}[t]
\caption{Error (less is better) for different levels of \textsc{Simple} noise.}
\label{tbl:NoiseLevel}
\begin{minipage}[b]{0.6\linewidth}
\setlength{\tabcolsep}{2.5pt}
\begin{tabular}{rrrrr}
& \multicolumn{3}{c}{Ours}\\
\cmidrule(lr{.75em}){2-4}
\multirow{2}{1 cm}{Noise Levels}&
No p.&
No c.&
Full&
Sup.
\\
&
\multicolumn{1}{c}{\textcolor{NoPriorColor}{\textbullet}}&
\multicolumn{1}{c}{\textcolor{NoColColor}{\textbullet}}&
\multicolumn{1}{c}{\textcolor{FullColor}{\textbullet}}&
\multicolumn{1}{c}{\textcolor{SupervisedColor}{\textbullet}}
\\
\toprule
1.5\,\% &.734 & .698 & .691 & .695\\
1.0\,\% & .578 & .534 & .525 & .515\\
0.5\,\% & .435 & .411 & .408 & .426\\
\bottomrule
\end{tabular}%
\end{minipage}%
\begin{minipage}[b]{0.4\linewidth}
\includegraphics[width=3.25cm]{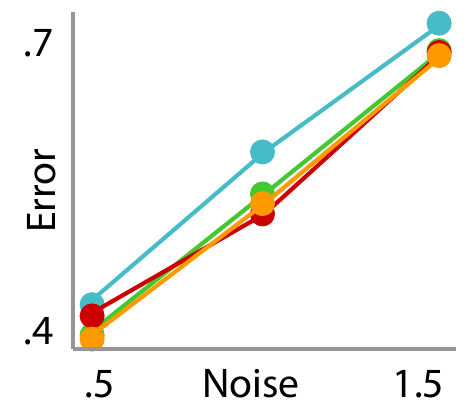}%
\vspace{-1.75cm}
\end{minipage}
\end{table}

While we have studied the average over three levels of noise in the previous plots, \refTbl{NoiseLevel} looks into the scalability with noise levels in units of scene diameter percentages. We find that error is increasing as expected with noise, but all methods do so in a similar fashion. In two cases, we win over supervised, in one case supervised wins, resulting in the improved average reported above.


\paragraph{Denoising performance.}

While we have studied \textsc{Simple} Gaussian noise, we now relax this assumption and explore \textsc{Advanced} simulated scanner noise generated as explained in \refSec{Setup}.
Contrary to real scanned data, it has the benefit that the ground truth is known.

\begin{table}[b]
\caption{Error (less is better) per method on \textsc{Advanced} noise.}
\label{tbl:NoiseType}
\begin{minipage}[b]{0.72\linewidth}
\setlength{\tabcolsep}{3.0pt}
\begin{tabular}{rrrrrr}
 & & \multicolumn{3}{c}{Ours} & \\
\cmidrule(lr{.75em}){3-5}
Mean&
Bilat.&
No p.&
No c.&
Full&
Sup.
\\
\multicolumn{1}{c}{\textcolor{MeanColor}{\textbullet}}&
\multicolumn{1}{c}{\textcolor{BilateralColor}{\textbullet}}&
\multicolumn{1}{c}{\textcolor{NoPriorColor}{\textbullet}}&
\multicolumn{1}{c}{\textcolor{NoColColor}{\textbullet}}&
\multicolumn{1}{c}{\textcolor{FullColor}{\textbullet}}&
\multicolumn{1}{c}{\textcolor{SupervisedColor}{\textbullet}}
\\
\toprule
.378 & .362 & .393 & .359 & .356 & .329\\
\bottomrule
\end{tabular}%
\end{minipage}%
\begin{minipage}[b]{0.28\linewidth}
\includegraphics[width=2cm, height=2cm]{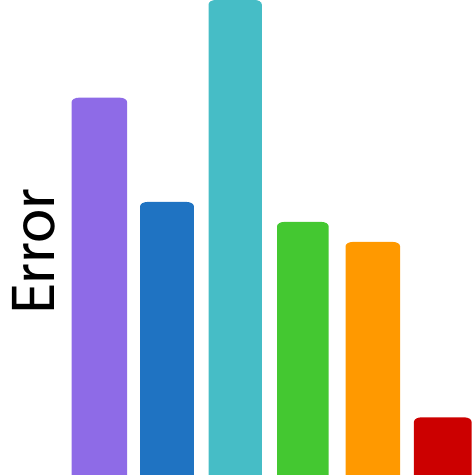}%
\vspace{-1.0cm}
\end{minipage}
\end{table}

\refTbl{NoiseType} shows the error of different methods for this type of noise.
We see, that in this case, our full method (orange) performs better than any other unsupervised method, such as mean or bilateral (violet and blue).
A supervised method can perform better than other methods for this noise at the same amount of training data input, $12\,$million points.
Finally, we also see that ablations without the suggested prior (cyan and green) have a higher error, indicating the priors are equally relevant for this type of noise, too.


\paragraph{Upgrading}
Notably, we can upgrade any supervised denoiser in a code-transparent, architecture-agnostic fashion to become unsupervised.
Consider a supervised denoiser, which takes clean-noisy pairs instead of noisy ones only, as we do.
To apply our method, all we do is to resample the point using our spatial and / or color prior, and ``pretend'' this to be the clean point cloud.

\begin{wraptable}{r}{3.cm}
\setlength{\tabcolsep}{1.5pt}
\centering
\label{tbl:FID}
\small
\begin{tabular}{rccc}
\toprule
&
\multicolumn{2}{c}{{PCNet \cite{rakotosaona2019pointclean}}}
\\
\cmidrule{2-3}
Noisy&{Sup.}&{Our}&
\\
\midrule
4.54&1.36&{\textbf{1.34}}\\
\bottomrule
\end{tabular}
\end{wraptable}
We have done so for PointCleanNet \cite{rakotosaona2019pointclean} and evaluated on their dataset.
We see even without modifying their architecture or supervision, we still slightly outperform theirs.

\mysubsection{Qualitative Results}{Qualitative}
Here, we repeat the above experiments, on real world noisy point clouds from a Mobile Laser Scanning setup based on a Velodyne HDL32 3D scanner. We used the Paris-rue-Madame data set~\cite{Serna2014ParisrueMadameD} which is composed of $20\,$million points. We subdivide the model into parts of ca. $150\,$K points each, resulting in $17\,$million points used during training and $3\,$million points for testing. Note that in this setting, noise is part of the data and does not need to be simulated. Furthermore, and most important, no clean ground truth is available. Consequentially, the error cannot be quantified and we need to rely on human judgment.

We see in \refFig{RealData}, how our method removes the noise  and produces a clean point cloud without shrinking, with uniform sampling as well as details. We cannot apply a visualization of the error, as the ground truth is unknown. We instead provided point cloud renderings by representing the point clouds as a mesh of spheres with shading.

\mysubsection{Ablation}{Ablation}

\paragraph{No prior}
When not using the prior in space (green), the denoiser learned across different types of noise (\refTbl{Performance}), magnitudes of noise (\refTbl{NoiseLevel}) and amounts of training data (\refTbl{Scalability}) is consistently worse and not much better than Gaussian or bilateral.
This indicates it is essential.

\paragraph{No appearance}
Making use of appearance consistently improves the outcome across the aforementioned three axes of variations in \refTbl{Performance} (and \refTbl{NoiseType}), \refTbl{Scalability} and \refTbl{NoiseLevel}, either taking the quality beyond supervision or very close to it.


\paragraph{Effect of color}
\refFig{CubesColor} shows a sharp edge with two different colors to be denoised.
Including the color, slightly reduces the error (less high-error yellow, more blue).
\myfigure{CubesColor}{Including and not including color in the prior.}{t}

\mysection{Conclusions}{Conclusion}
We have presented an unsupervised learning method to denoise 3D point clouds without needing access to clean examples, and not even noisy pairs. This allows the method to scale with natural data instead of clean CAD models decorated with synthetic noise. Our achievements were enabled by a network that maps the point cloud to itself in combination with a spatial locality and a bilateral appearance prior.
Using appearance in the prior is optional, but can improve the result, without even being input to the network, neither at test nor at training time.
Denoising with color as input, as well as joint denoising of color and position, remains future work.
Our results indicate we can outperform supervised methods, even with the same number of training examples.

\vspace*{-0.15cm}\paragraph{Acknowledgements}
\small This work was partially funded by the Deutsche Forschungsgemeinschaft (DFG), grant RO 3408/2-1 (ProLint), and the Federal Ministry for Economic Affairs and Energy (BMWi), grant ZF4483101ED7 (VRReconstruct). We acknowledge Gloria Fackelmann for the supplementary video narration.

{\small
\bibliographystyle{ieee_fullname}
\bibliography{main}
}

\clearpage

\begin{strip}
\centering
{\Large \textbf{Supplementary material}}
\vspace{1cm}
\end{strip}

\appendix

\mysection{Error distribution}{ErrorDistribution}

\begin{figure}[b]
\centering
\begin{tabular}{m{3.8cm} m{3.8cm}} 
    \includegraphics*[width =\linewidth]{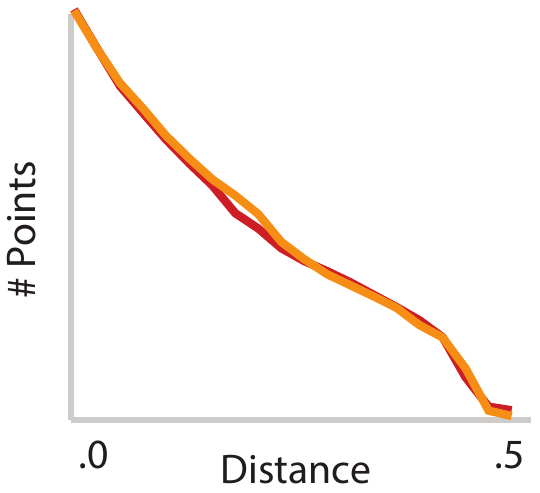} &
    \includegraphics*[width =\linewidth]{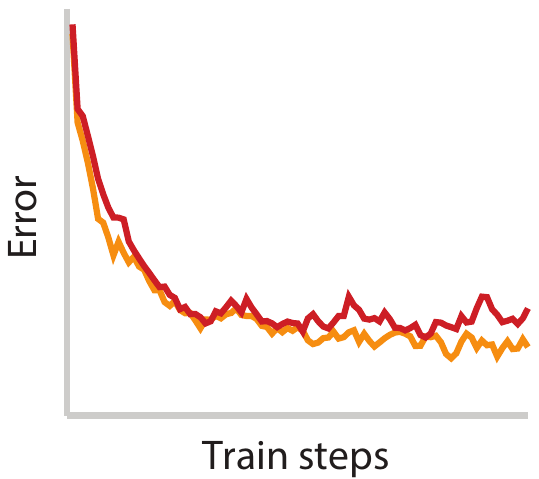} \\
    \multicolumn{1}{c}{Error histogram.}&
    \multicolumn{1}{c}{Eval. loss during training.}
\end{tabular}
\vspace{.15cm} 
\caption{\textbf{Ours (\textsc{FULL})} \textcolor{FullColor}{\textbullet} vs \textbf{Supervised} \textcolor{SupervisedColor}{\textbullet}}
\label{fig:SupHistogramLoss}
\end{figure}

We studied the distribution of points wrt. the distance to the underlying surface in our test dataset.
\refFig{SupHistogramLoss} (left) presents a histogram with the results of such study.
Here we can see how the distribution of points has its maximum at distance $0.0$ from the real surface and decreases with distance, following the same distribution as in the supervised setting.

\mysection{Convergence}{Convergence}

\refFig{SupHistogramLoss} (right) illustrates the evaluation error during training for two networks trained on the smallest dataset ($.5$\,Million points).
One network is trained with our algorithm (\textsc{Full}) and the other one with supervised training.
We can see that in both cases the loss decreases during training similarly.
However, our algorithm converges to a lower error than supervised learning.
With more training data, this gap is reduced until the curves cross.

\mysection{Prior Kernel}{ChoiceKernel}

We experimented with different kernels for our prior: Gaussian ($\kernel_{g}$), Wendland ($\kernel_{w}$), and Inverse Multi-Quadric ($\kernel_{i}$).
All kernels were adjusted to fit in the range $[-1, 1]$:

\begin{align}
\kernel_{g}(\offset)&={}\frac{1}{\sigma\sqrt{2\pi}}\exp\left(-\frac{||\proximityMatrix\offset||_2^2}{2\sigma^2}\right), \nonumber \\
\kernel_{w}(\offset)&={}(1-||\proximityMatrix\offset||_2^2)^4(1+4||\proximityMatrix\offset||_2^2),\nonumber \\
\kernel_{i}(\offset)&={}\frac{1}{\sqrt{1+(5||\proximityMatrix\offset||_2^2)^2}}, \nonumber
\end{align}

\noindent where $\proximityMatrix=\operatorname{diag}(\proximityMatrixElement)$ and $\proximityMatrixElement=1/\pathcRadiusScale\pathcRadius$. We performed an ablation study for different values of $\pathcRadiusScale$ and for each  kernel independently. 
For the Gaussian kernel we used $\pathcRadiusScale_1=.3$, $\pathcRadiusScale_2=.5$, and $\pathcRadiusScale_3=.7$, and for the Wendland and Inverse Multi-Quadric kernel we used $\pathcRadiusScale_1=.8$, $\pathcRadiusScale_2=1.$, and $\pathcRadiusScale_3=1.2$. \refTbl{supkernel} shows that the best performance was obtained by the Gaussian kernel.

\begin{table}[t]
\centering
\caption{Error obtained for different functions used as our prior.}
\label{tbl:supkernel}
\begin{minipage}[b]{0.7\linewidth}
\begin{tabular}{rrrr}
Prior&
$\pathcRadiusScale_1$&
$\pathcRadiusScale_2$&
$\pathcRadiusScale_3$
\\
\toprule
Gauss. ($\kernel_{g}$) \textcolor{FullColor}{\textbullet}& .567 & .548 & .556\\
Wedland ($\kernel_{w}$) \textcolor{SupervisedColor}{\textbullet}& .595 & .560 &.561\\
Inv. MQ ($\kernel_{i}$) \textcolor{NoColColor}{\textbullet}& .577 & .577 & .582\\
\bottomrule
\end{tabular}%
\end{minipage}%
\begin{minipage}[b]{0.3\linewidth}
\includegraphics[width=2.5cm, height=2cm]{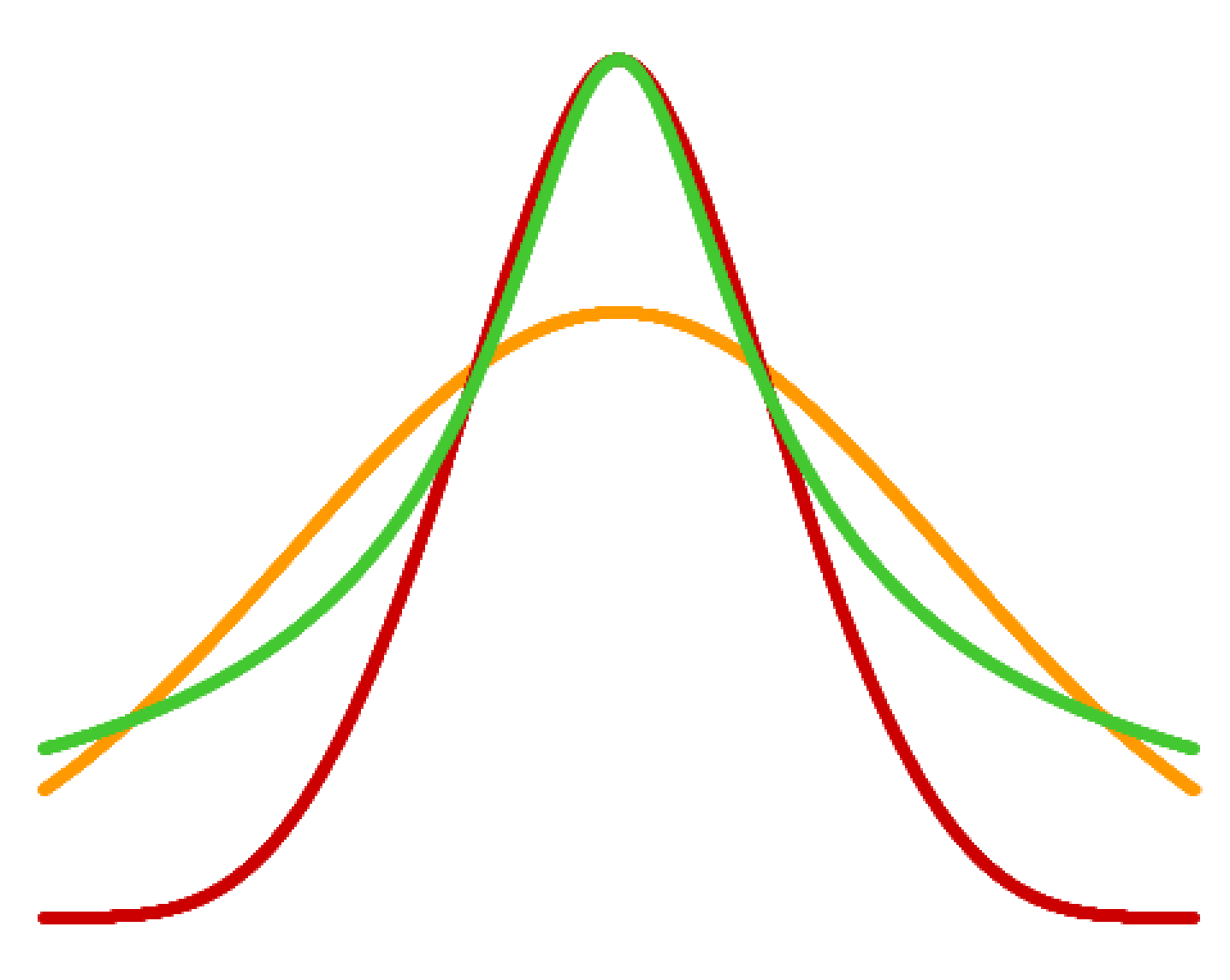}%
\vspace{-1.0cm}
\end{minipage}
\end{table}

\mysection{Noise levels}{NoiseLevels}

\begin{wraptable}{r}{3.5cm}
\setlength{\tabcolsep}{2pt}
\centering
\label{tbl:FID}
\small
\begin{tabular}{crrr}
\toprule
&
\multicolumn{3}{c}{Test}\\
\cmidrule(lr{.75em}){2-4}
Train&
0.5\%&
1.0\%&
1.5\%\\
\midrule
0.5\%&.435&.576&.738\\
1.0\%&.431&.551&.708\\
1.5\%&.467&.594&.741\\
All&.411&.534&.698\\
\bottomrule
\end{tabular}
\end{wraptable}

Tbl. 3 on the paper tests one neural network trained for all noise levels at the same time. In this experiment, we have trained our network using only one level of noise and report its test error for all levels. We can see from the table, that performance drops only marginally.

\mysection{Iterative refinement}{IterativeRefinement}

\begin{figure*}[!t]
\centering
\includegraphics*[width =\linewidth]{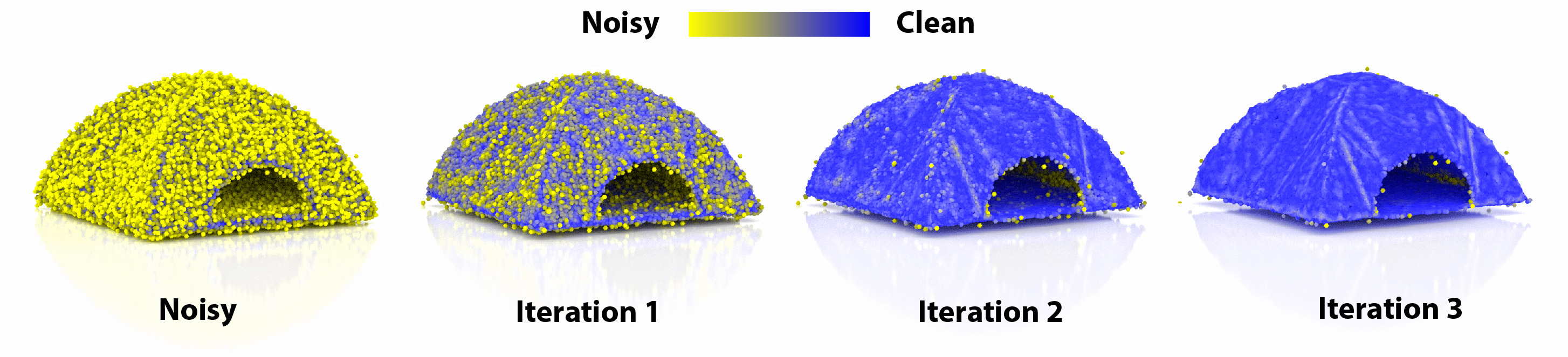}
\caption{Resulting point cloud of applying our network several times iteratively on a noisy point cloud.}
\label{fig:Iterations}
\end{figure*}

Since our approximation of the surface during training is not exact, the results improve by applying our network several times iteratively. \refFig{Iterations} presents a visual encoding of the distance of each point to the real surface after applying each denoising step.

\begin{figure*}[!t]
\centering
\includegraphics*[width =\linewidth]{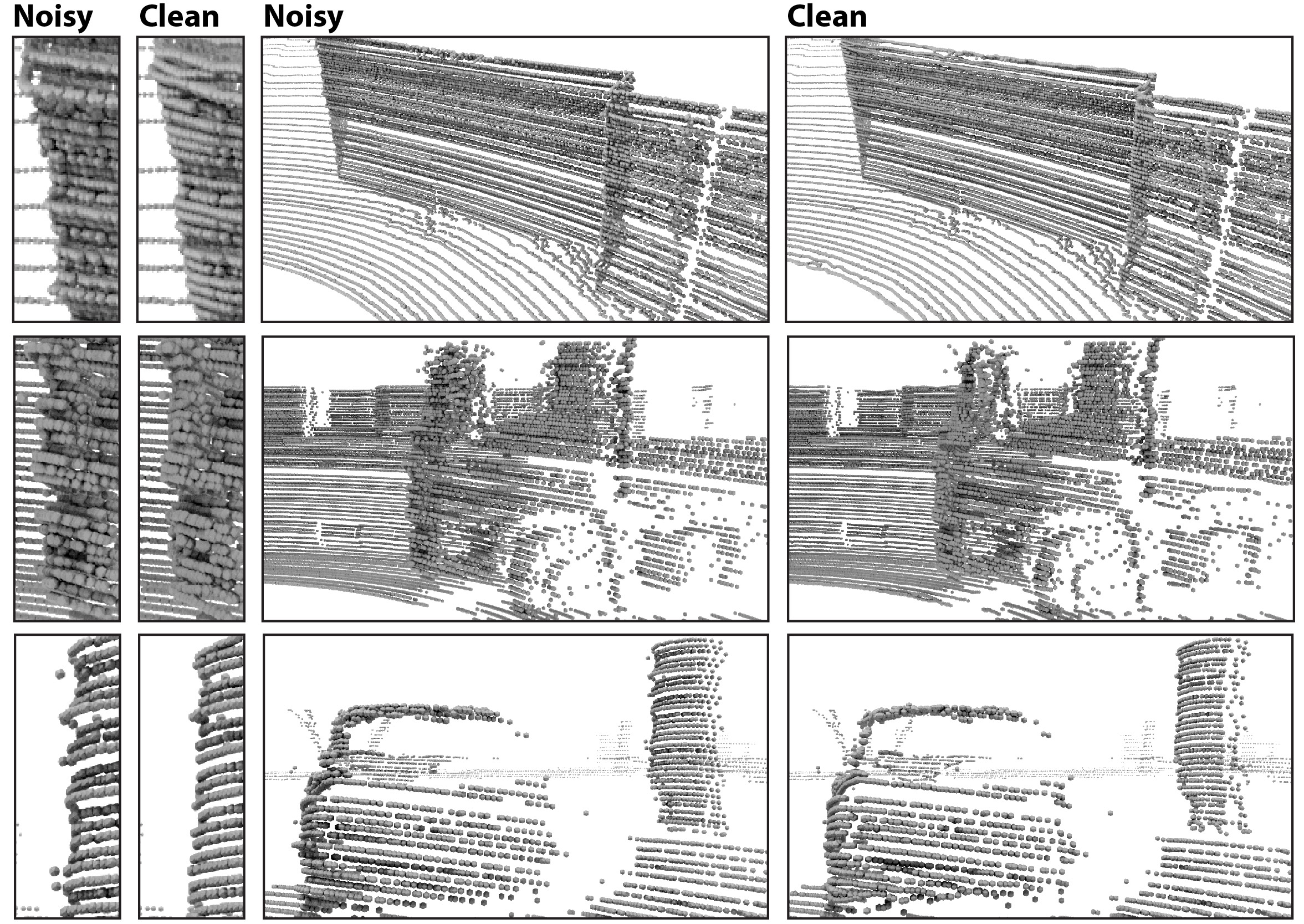}
\caption{Qualitative results of our network trained on the Kitti dataset~\cite{Geiger2013IJRR}. Note how our network is able to remove the noise in sparse point clouds as the one illustrated in the bottom image.}
\label{fig:SupKitti}
\end{figure*}

\mysection{Additional Qualitative Results}{MoreQualitativeResults}

We trained our network with the Kitti dataset~\cite{Geiger2013IJRR} in order to evaluate the robustness of our method with spare point clouds. \refFig{SupKitti} presents some qualitative results. We can see that our network is able to remove the noise successfully. 

\begin{figure*}[!t]
\centering
\includegraphics*[width =\linewidth]{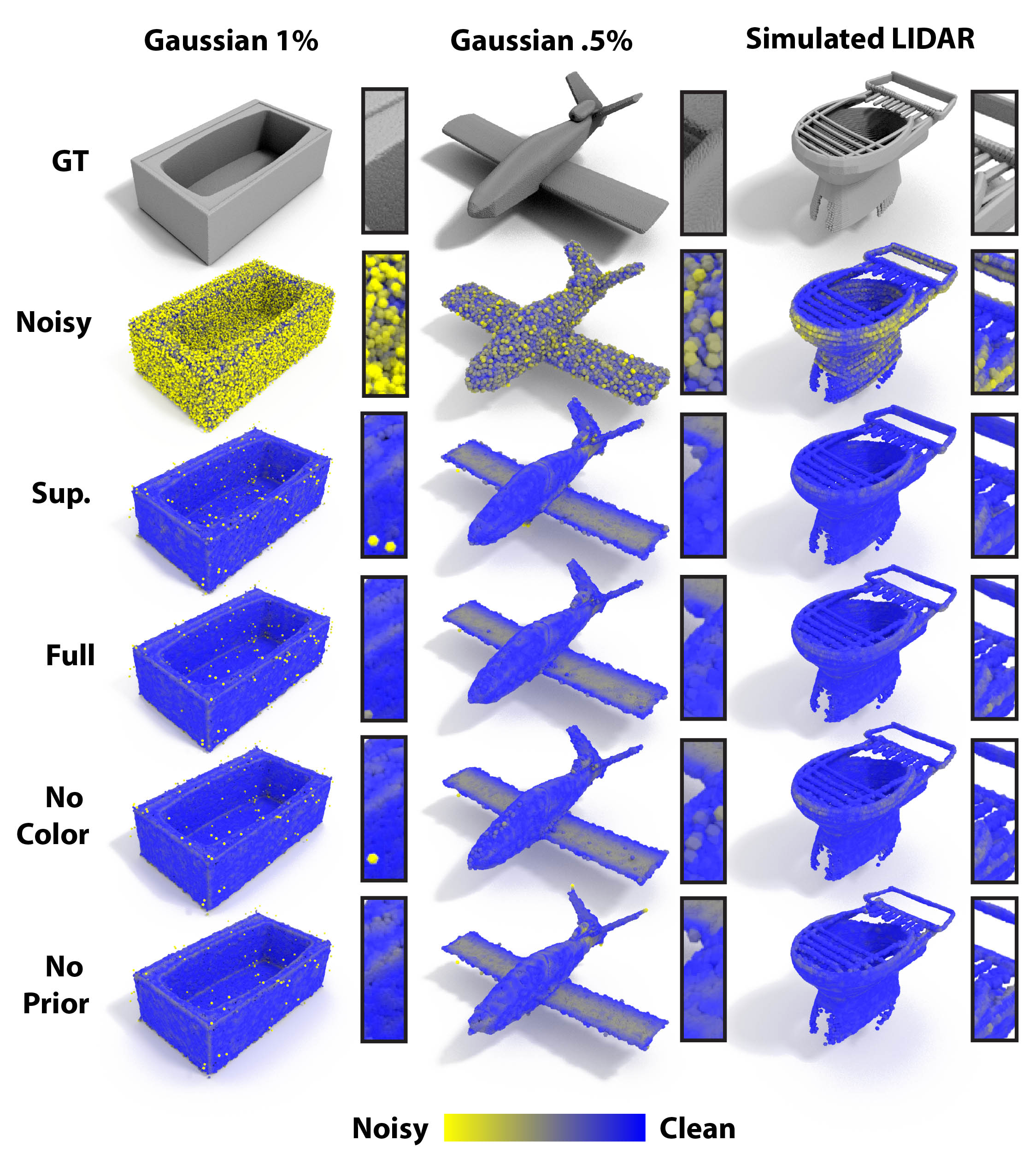}
\caption{Visual encoding of the distance from each point to the underlying ground truth surface. Here blue indicates zero distance to the surface and yellow $2\%$ of the diagonal of the model. We can see that for the two simulated datasets with different levels of noise, our algorithm achieves (\textbf{Full}) almost the same quality as a network trained with supervised data (\textbf{Sup.}). Moreover, we evaluated two variants of our training procedure: a prior without color information (\textbf{No Color}), and \textbf{No Prior}.}
\label{fig:SupSimulated}
\end{figure*}

\begin{figure*}[!t]
\centering
\includegraphics*[width =\linewidth]{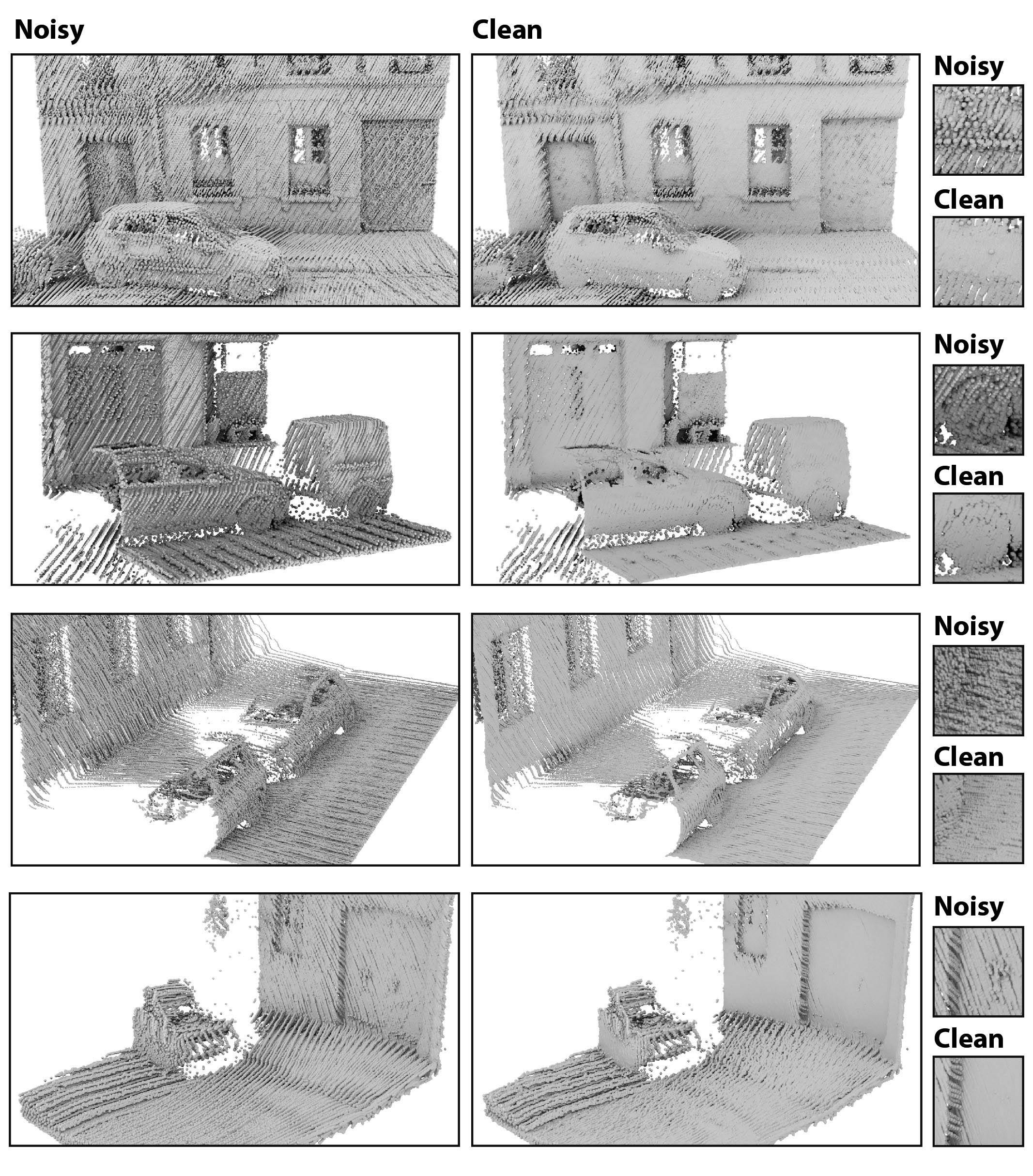}
\caption{Comparison of our denoising algorithm \textbf{(right)} with the noisy scanned data \textbf{(left)} from the Paris-rue-Madame Database ~\cite{Serna2014ParisrueMadameD}. The number of points for each model are, from top to bottom, $800$\,K, $450$\,K, $450$\,K, and $375$\,K points. Our network is able to process each model in parallel in a workstation with a single Nvidia RTX 2080.}
\label{fig:SupRue}
\end{figure*}

Moreover, \refFig{SupSimulated}, and \refFig{SupRue} provide more qualitative results on all the datasets used in the evaluation.
In \refFig{SupSimulated} we can see how our algorithm obtains a quality similar to the network trained with supervised data.
\refFig{SupRue} illustrates the result of applying our method to models from the Paris-rue-Madame Database~\cite{Serna2014ParisrueMadameD} composed of $375$\,K-$800$\,K points.

\end{document}

%% file: tr-commands.tex
\usepackage{soul}
\usepackage{amsmath}
\usepackage{amssymb}
\usepackage{wrapfig}

\def\figurePath{}
\def\myfigure#1#2#3{\begin{figure}[#3]\centering\includegraphics*[width = \linewidth]{\figurePath#1}\caption{#2}\label{fig:#1}\end{figure}}

\def\mycfigure#1#2{\begin{figure*}[t]\centering\includegraphics*[clip, width = \linewidth]{\figurePath#1}\caption{#2}\label{fig:#1}\end{figure*}}

\renewcommand{\eg}{e.\,g., }
\renewcommand{\ie}{i.\,e., }
\renewcommand{\etal}{et~al.\ }

\newcommand{\argmin}[1]{\underset{#1}{\operatorname{arg\,min}}}

\newcommand{\refSec}[1]{Sec.~\ref{sec:#1}}
\newcommand{\refFig}[1]{Fig.~\ref{fig:#1}}
\newcommand{\refEq}[1]{Eq.~\ref{eq:#1}}
\newcommand{\refTbl}[1]{Tbl.~\ref{tbl:#1}}
\newcommand{\refAlg}[1]{Alg.~\ref{alg:#1}}

\soulregister\ref7
\soulregister\cite7
\soulregister\refFig7
\soulregister\cite7
\soulregister\ref7
\soulregister\pageref7
\soulregister\shortcite7
\soulregister\eg0
\soulregister\ie0
\soulregister\etal0

\DeclareGraphicsExtensions{.png,.jpg,.pdf,.ai,.psd}
\newcommand{\mysection}[2]{\section{#1}\label{sec:#2}}
\newcommand{\mysubsection}[2]{\subsection{#1}\label{sec:#2}}